%% file: iccp21_template.tex
\documentclass[10pt,journal,compsoc]{IEEEtran}
\newif\ifpeerreview

\peerreviewfalse

\usepackage[nocompress]{cite}
\usepackage{url}
\usepackage{amsmath,amssymb,graphicx}
\usepackage{multirow}
\usepackage{tabularx}
\usepackage{bm}
\usepackage{color, colortbl}
\usepackage[T1]{fontenc}\usepackage{hyperref}
\hypersetup{
    colorlinks=true,
    linkcolor=blue,
    filecolor=magenta,      
    urlcolor=magenta,
}
\newcommand{\highlighttext}[1] {{#1}}

\usepackage{lipsum} 

\usepackage[switch]{lineno}

\newcommand{\paperID}{13}

\title{Multi-Stage Raw Video Denoising with Adversarial Loss and Gradient Mask}

\author{Avinash~Paliwal, Libing~Zeng and~Nima~Khademi~Kalantari
\IEEEcompsocitemizethanks{\IEEEcompsocthanksitem A. Paliwal, L. Zeng and N.K. Kalantari are with the Department of Computer Science and Engineering, Texas A\&M University, College Station, TX 77843. E-mail: \{avinashpaliwal, libingzeng, nimak\}@tamu.edu}\protect\\
}

\begin{document}

\IEEEtitleabstractindextext{%
\begin{abstract}
In this paper, we propose a learning-based approach for denoising raw videos captured under low lighting conditions. We propose to do this by first explicitly aligning the neighboring frames to the current frame using a convolutional neural network (CNN). We then fuse the registered frames using another CNN to obtain the final denoised frame. To avoid directly aligning the temporally distant frames, we perform the two processes of alignment and fusion in multiple stages. Specifically, at each stage, we perform the denoising process on three consecutive input frames to generate the intermediate denoised frames which are then passed as the input to the next stage. By performing the process in multiple stages, we can effectively utilize the information of neighboring frames without directly aligning the temporally distant frames. We train our multi-stage system using an adversarial loss with a conditional discriminator. Specifically, we condition the discriminator on a soft gradient mask to prevent introducing high-frequency artifacts in smooth regions. We show that our system is able to produce temporally coherent videos with realistic details. Furthermore, we demonstrate through extensive experiments that our approach outperforms state-of-the-art image and video denoising methods both numerically and visually.
\end{abstract}

}

\ifpeerreview
\linenumbers \linenumbersep 15pt\relax 
\author{Paper ID \paperID\IEEEcompsocitemizethanks{\IEEEcompsocthanksitem This paper is under review for ICCP 2021 and the PAMI special issue on computational photography. Do not distribute.}}
\markboth{Anonymous ICCP 2021 submission ID \paperID}%
{}
\fi
\maketitle
\thispagestyle{empty}
\input{Introduction.tex}
\input{RelatedWork.tex}
\input{Algorithm.tex}

\input{Results.tex}

\input{Conclusion.tex}

\ifpeerreview \else
\section*{\highlighttext{Acknowledgments}}
\highlighttext{The authors would like to thank the reviewers for their comments and suggestions. The synthetic \textit{test} dataset was collected from YouTube channels {\em Video Library - No copyright Footage}, {\em Le Monde en Vidéo} and {\em Underway}, all under Creative Commons (CC) license. }
\bibliographystyle{IEEEtran}
\bibliography{egbib}

\ifpeerreview \else





\vspace{-0.25in}
\begin{IEEEbiography}[{\includegraphics[width=1in,height=1.25in,clip,keepaspectratio]{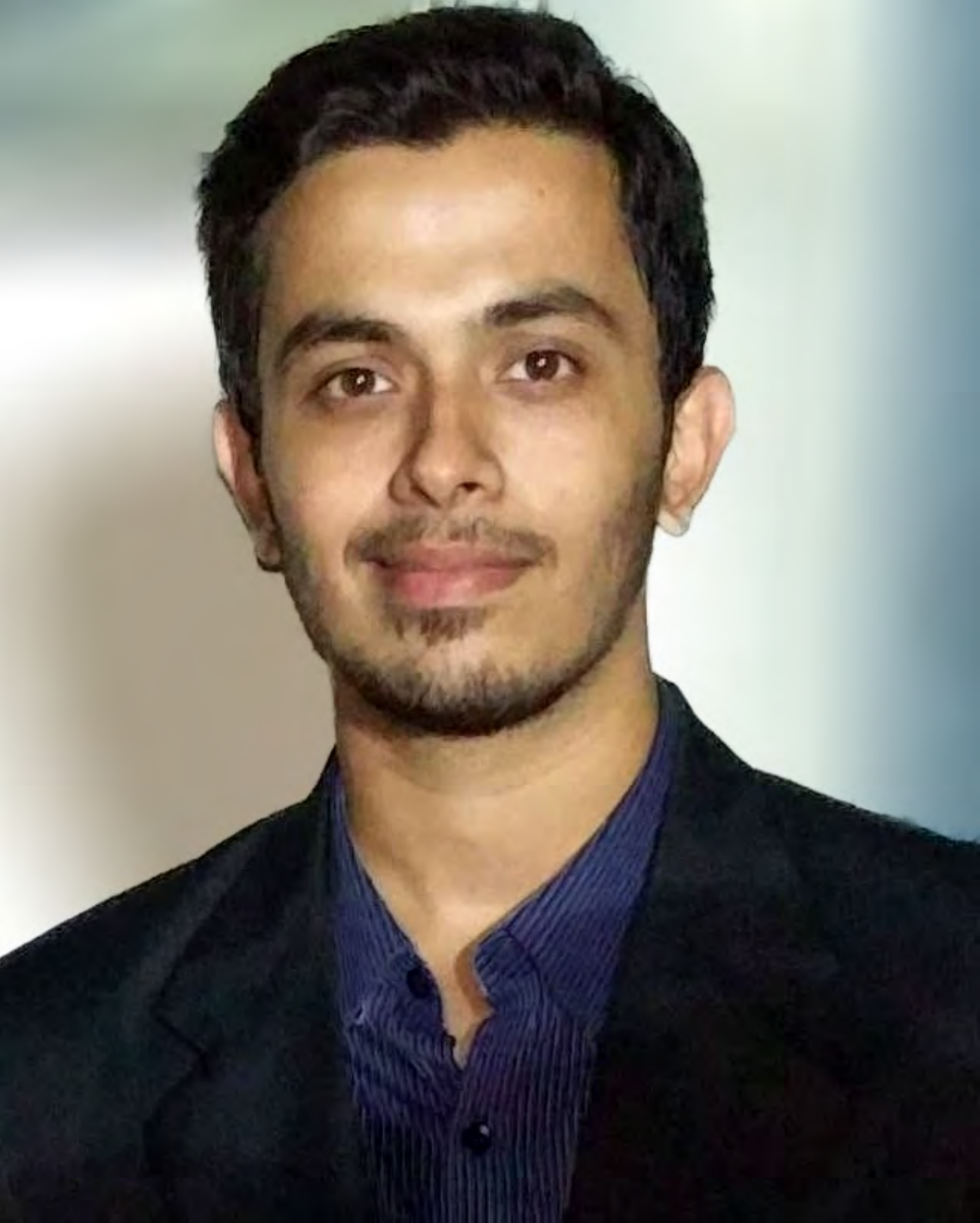}}]{Avinash Paliwal}
is a Ph.D. student at the Computer Science and Engineering Department of Texas A\&M University, College Station, TX, USA. He received his B.Tech. degree in Electronics and Communication Engineering from Visvesvaraya National Institute Of Technology, Nagpur, India. His research interests are in computational photography, computer vision and computer graphics.
\end{IEEEbiography}
\vfill
\vspace{-0.5in}
\begin{IEEEbiography}[{\includegraphics[width=1in,height=1.25in,clip,keepaspectratio]{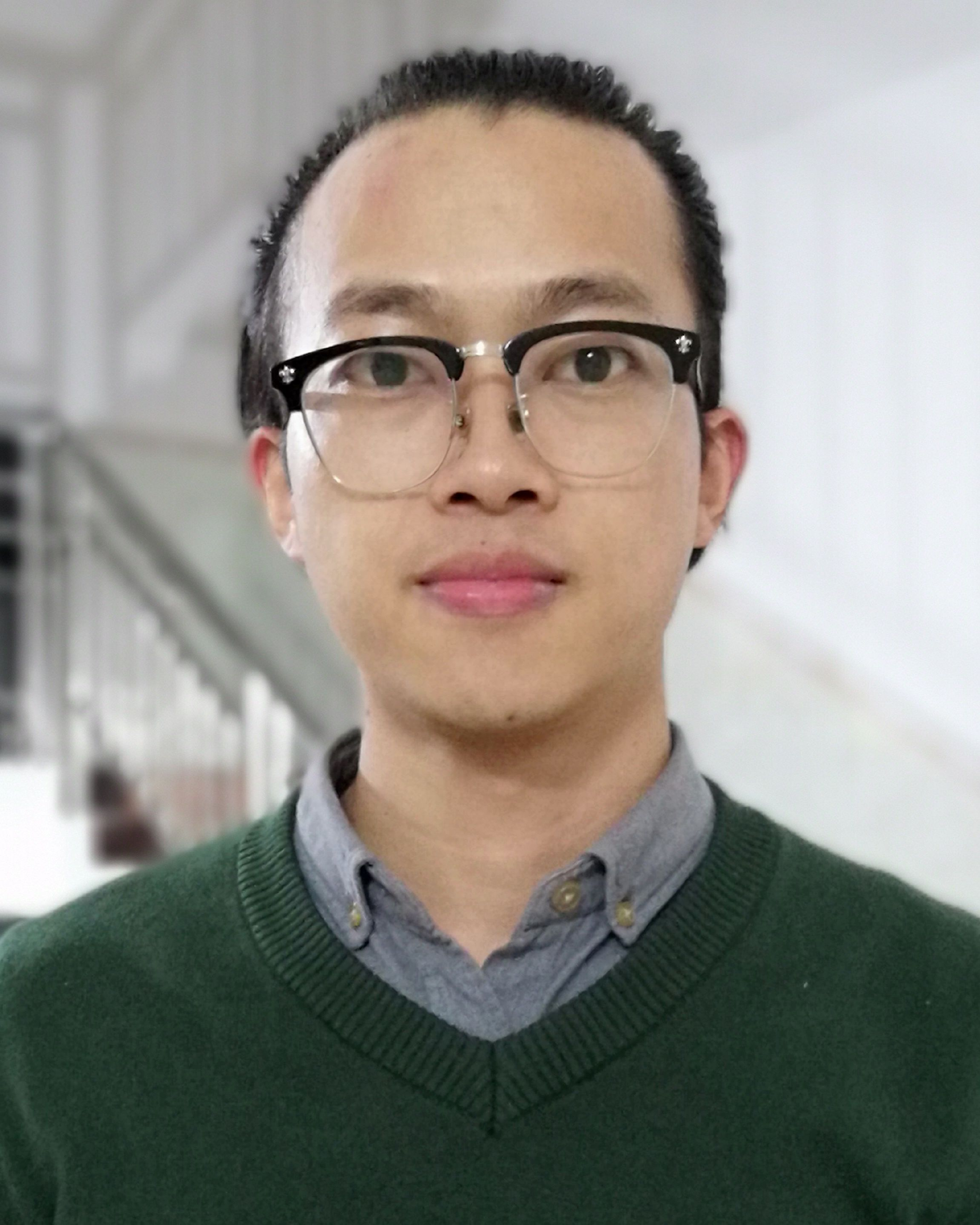}}]{Libing Zeng}
is a Ph.D. student at the Computer Science and Engineering Department of Texas A\&M University, College Station, TX, USA. He received his Bachelors degree in Electronic Information Engineering from Hunan University, China. His research interests are in computational photography, computer vision and deep learning.
\end{IEEEbiography}
\vfill
\vspace{-0.5in}
\begin{IEEEbiography}[{\includegraphics[width=1in,height=1.25in,clip,keepaspectratio]{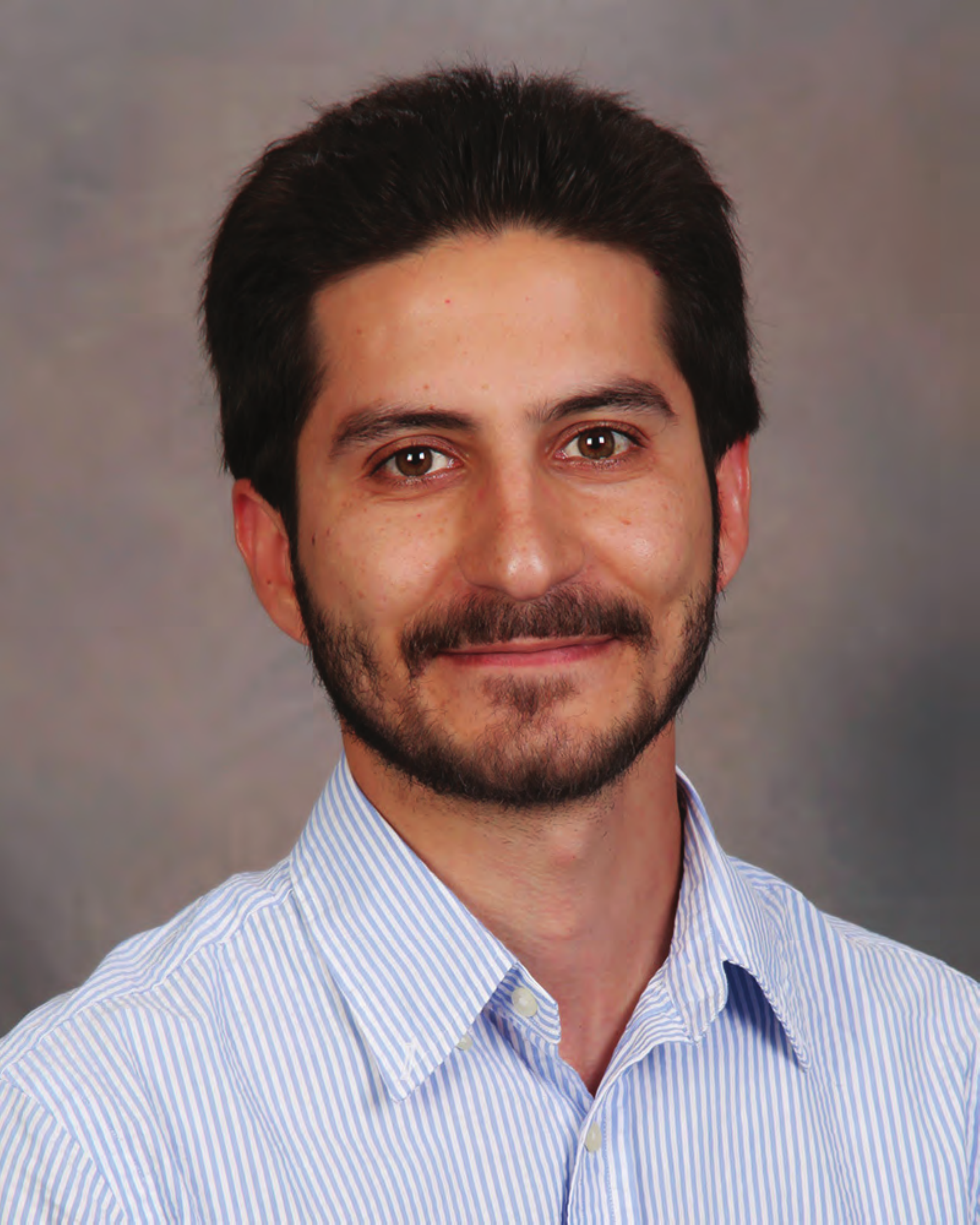}}]{Nima Khademi Kalantari}
is an assistant professor in the Computer Science and Engineering department at Texas A\&M University. He received his doctoral degree in Electrical and Computer Engineering from the University of California, Santa Barbara. Prior to joining Texas A\&M, he was a postdoc in the Computer Science and Engineering department at the University of California, San Diego. His research interests are in computer graphics with an emphasis on computational photography and rendering. In recent years, he has focused on developing deep learning techniques for image synthesis in these two fields. He currently serves as an Associate Editor for ACM Transactions on Graphics.
\end{IEEEbiography}

\fi

\end{document}

%% file: Introduction.tex
\IEEEraisesectionheading{
  \section{Introduction}\label{sec:introduction}
}
\IEEEPARstart{C}{apturing} videos of dynamic scenes in low-light environments, such as at night or indoors, is inevitable.
Videography in these situations is challenging because the number of photons arriving at camera sensors is low,  resulting in severe noise in the captured videos.
In addition to reduction in visual quality, the existence of noise negatively affects the subsequent video processing tasks.
Hence, video denoising plays an important role in video processing, especially under low lighting conditions.

With decades of studies in image denoising, a large number number of non-learning \cite{Rudin1992, Portilla2003, Elad2006, Dabov2007, Mairal2009, Heide2014, Gu2014, Schmidt2014} and learning-based techniques \cite{Jain2008Neural, Burger2012Neural, Agostinelli2013DenoisingDeep, Zhang2017DenoisingCNN, Chen2017Trainable, Divakar2017ImageDenoisingGAN, Chen2018ImageDenoisingGAN} have been developed to remove additive white Gaussian noise (AWGN).
However, the noise in final sRGB images is transformed from raw images through an image signal processor (ISP) pipeline and is more complicated than AWGN.
Therefore, the performance of these approaches on real-world images is limited.
The most recent approaches \cite{Chen2018aSID, brooks2019unprocessing, Zamir2020CycleISP} address this issue by using more sophisticated noise models and performing the denoising on raw images.

While these approaches can successfully denoise a single image, they are not effective at denoising videos, producing results with severe flickering artifacts. 
This is mainly because they do not take the information of neighboring frames into consideration.
A few approaches \cite{Maggioni2012, Chen2016, Xue2017, Claus2019, Tassano2019DVDNETAF, Tassano2020FastDVDnet} propose to tackle this application by considering the neighboring frames when removing AWGN from an sRGB video frame, but their performance on real-world videos is sub-optimal.

Most recently, a couple of approaches address this problem by performing the denoising on raw video frames.
Chen et al. \cite{Chen2019SMD} propose to train a single frame denoising network on static videos and enforce the results of consecutive frames to be similar.
However, their method does not utilize the information of adjacent frames and, thus, severe flickering artifacts still occur in videos with large motions.
Yue et al. \cite{Yue2020} incorporate the information of neighboring frames by implicitly aligning their features using deformable convolutions \cite{Dai2017Deformable}. 
However, this alignment component cannot be directly supervised as it registers the feature maps implicitly.
Therefore, their approach is not able to effectively use the information of neighboring frames, producing results with blurriness and flickering.

To address this problem, we propose a novel learning-based approach that denoises each frame by explicitly aligning  the neighboring frames.
Specifically, we first align the neighboring frames to the current frame and then fuse the registered frames to obtain the final denoised frame. 
Furthermore, since aligning the temporally distant neighboring frames is difficult, we propose to perform the two processes of alignment and fusion in multiple stages. 
At each stage, we use three consecutive frames to denoise the middle frame. By repeating this process in each stage, we can effectively utilize the information of multiple neighboring frames to produce each denoised frame. 
With our multi-stage approach, the alignment is always performed between the immediate neighboring and current frames, thereby alleviating the difficulty of aligning temporally distant frames.

We train our multi-stage system in an end to end manner on a large set of synthetically generated noisy and clean videos. 
To encourage the network to produce sharp frames with realistic details, we propose to train the network using an adversarial loss function. 
Furthermore, to prevent the network from producing high-frequency artifacts in the smooth regions, we condition the discriminator on a soft gradient mask.

We demonstrate the superiority of our framework through comprehensive comparisons against the state-of-the-art denoising methods on a variety of synthetic and real scenes. Moreover, we show the effectiveness of our design choices through extensive ablation studies.

Our contributions are three-fold:
\begin{itemize}
    \item A multi-stage flow-based video denoiser which can effectively utilize the information of multiple neighboring frames to help denoise the center frame.
    \item We propose to train our network in an adversarial manner using a discriminator which is conditioned on a soft gradient mask. 
    Our system produces results with realistic details without introducing high-frequency artifacts in the smooth regions.
    \item We demonstrate that our approach outperforms state-of-the-art image and video denoising methods on a variety of real and synthetic test scenes.
\end{itemize}


%% file: RelatedWork.tex
\section{Related Work}

In this section, we briefly review the image and video denoising methods, including non-learning and learning-based approaches.
Since we use an adversarial loss to train our network, we also provide a brief overview of generative adversarial networks (GANs).

\subsection{Image Denoising}

The problem of image denoising plays an important role in image processing and computer vision and has been the subject of extensive research in the past.
Traditional approaches \cite{Rudin1992, Portilla2003, Buades2005NLM, Elad2006, Dabov2007, Mairal2009, Heide2014, Gu2014} reconstruct a clean image from a noisy image by taking advantage of either local statistics or self-similarity of the input image.
In recent years, learning-based algorithms \cite{Burger2012Neural, Agostinelli2013DenoisingDeep, Gharbi2016DDD, Zhang2017DenoisingCNN, Chen2017Trainable, UlyanovVL18, Ploetz2018NNNN, Zhang2018ISRUVDRCAN, Guo2019TCBD, brooks2019unprocessing} have significantly advanced the field by harnessing the power of deep neural networks.

However, the aforementioned approaches perform denoising on sRGB images for removing AWGN and are not able to generalize to real-world images.
To address this problem, Chen et al. \cite{Chen2018aSID} train a network on pairs of real noisy and clean raw images. 
Specifically, they capture the noisy image in low lighting conditions and obtain the ground truth clean image by capturing a long exposure image of the scene.
Guo et al. \cite{Guo2019TCBD} and Brooks et al. \cite{brooks2019unprocessing} train their networks on synthetic raw images generated by unprocessing existing sRGB images.
To generate the input, they accurately model the camera noise and add it to the synthetically generated raw images.
Although these methods demonstrate promising performance on image denoising, they are not able to effectively denoise videos as they do not exploit the information of neighboring frames.
Therefore, they are generally not able to produce temporally coherent results.

Several approaches propose to generate a denoised image from burst images~\cite{mildenhall2018kpn, Xia_2020_CVPR, Hasinoff_2016_BurstPh}. These approaches utilize the information of multiple images to reconstruct a single denoised image and could be potentially used for generating denoised videos. However, these burst photography algorithms are mainly designed for input images with small motion, while the motion in typical videos are generally large.

\subsection{Video Denoising}

Compared to image denoising, video denoising is a more challenging problem which is relatively under-explored.
Existing non-learning video denoising methods \cite{Maggioni2012, Lebrun2013, Ghoniem2010NVDSIUDRG, Han2012EVDBDNM, Gui2017VDULRTD, buades2016patch} generate results by grouping similar patches and then jointly filtering them.
In recent years, learning-based video denoising methods have demonstrated promising results.
Chen et al. \cite{Chen2016} use a simple fully connected RNN to perform video denoising on sRGB frames.
Xue et al. \cite{Xue2017} propose to handle this application by aggregating information from registered frames using learned task-oriented flows. 
We also explicitly register the neighboring frames, but we do so through a multi-stage system to be able to effectively align the temporally distant frames.
Claus et al. \cite{Claus2019} propose a CNN video denoiser using a combination of spatial and temporal filtering without prior knowledge of the noise distribution.
Tassano et al. \cite{Tassano2019DVDNETAF, Tassano2020FastDVDnet} propose a two-step cascaded architecture without explicit motion estimation.
We also propose to perform the denoising in multiple stages, but explicitly align the neighboring images to effectively utilize their information.
Moreover, all the denoising blocks in our multi-stage denoiser use shared weights, reducing the number of trainable parameters in our system.

Unfortunately, these approaches operate on the sRGB images and are specifically designed to remove AWGN from sRGB images, reducing their performance on real-world videos.
Recently, a few methods propose to perform denoising on raw videos to be able to effectively denoise real-world videos.
Chen et al. \cite{Chen2019SMD} propose to train a single frame denoising network on low-light static videos with the long exposure image of the same scene as ground truth.
To achieve temporal coherency, they enforce the results of consecutive frames to be similar.
However, their method does not utilize the information of adjacent frames and, thus, severe flickering artifacts still occur in videos with large motions.
Yue et al. \cite{Yue2020} propose to use deformable convolutions \cite{Dai2017Deformable} to implicitly align features in neighboring frames.
However, directly supervising the alignment component is not possible as the feature maps are implicitly registered.
Therefore, their approach still produces results with blurriness and flickering due to the failure of effectively using the information of neighboring frames.

\subsection{Generative Adversarial Networks}

Since the introduction of generative adversarial network (GAN) by Goodfellow et al. \cite{Goodfellow2014}, it has been widely used to produce visually convincing images for computer vision tasks, such as image synthesis \cite{Brock2018, karras2019style, Zhang2018a, Tang2020}, image translation \cite{Tang2018, Tang2019}, and image super-resolution \cite{Ledig2017SRGAN, Bulat2018SRgAN, You2020CTSRGAN, Wang2019ESRGAN}.
More related to our application, Chen et al. \cite{Chen2018ImageDenoisingGAN} leverage a GAN to model the noise in real images.
GANs have also been used to remove noise from Monte Carlo rendered images \cite{Xu2019MCGAN}, tomography images \cite{Chen2020BioGAN}, and real images \cite{Zhong2019GAN}. 
In this paper, we use adversarial loss for video denoising and propose to condition the discriminator on a soft gradient to reduce the high-frequency artifacts in the smooth areas.

%% file: Algorithm.tex
\section{Algorithm}

The goal of our system is to denoise a real raw video sequence, while preserving the sharp features of the scene. To do this, we generate each denoised frame, $I^{d}_{t}$, by utilizing the information of $2N+1$ noisy consecutive frames, $I^{n}_{[t-N:t+N]}$, where $N>1$. We separate the raw bayer pattern of each input image into an image with four channels~\cite{brooks2019unprocessing, Zamir2020CycleISP, Yue2020}, as shown in Fig.~\ref{fig:denoiser}, before passing it to our denoiser. Once a denoised raw frame is obtained using our system, we generate the sRGB frame through the pretrained image signal processor (ISP) by Yue et al.~\cite{Yue2020}.

We divide the video denoising task into two stages of alignment and fusion, and model them using two convolutional neural networks (CNNs). During alignment, the neighboring frames are aligned with the reference frame. The aligned images are then fused to synthesize the denoised output during the fusion stage. To effectively utilize the information of the neighboring frames, we perform the denoising process
in multiple stages. We train our networks using an adversarial loss with a conditional discriminator to ensure producing high-quality videos with realistic details. The overview of our system is shown in Fig.~\ref{fig:multistage} and we discuss each stage in detail in the following sections.

\subsection{Multi-Stage Denoiser}
\begin{figure}
\includegraphics[width=\linewidth]{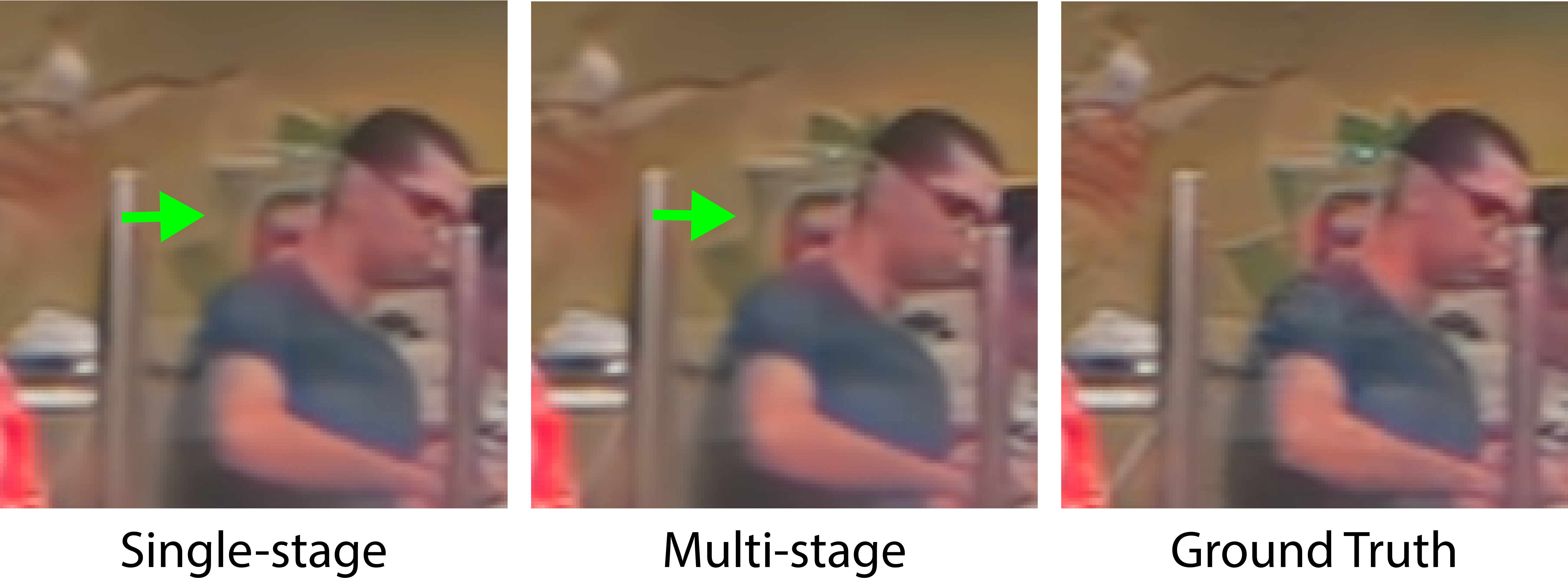}
\caption{Comparison of the single-stage approach against our multi-stage technique when 5 consecutive input frames are used.}
\label{fig:ablation}
\end{figure}

As discussed, we would like to generate the denoised frame, $I^d_t$, by first aligning the $2N$ neighboring frames to the central frame, $I^n_t$, and then fusing the $2N+1$ registered frames. However, as the temporal distance between the neighboring and reference frames increases, the alignment quality deteriorates due to complex large object and camera motion (see Fig.~\ref{fig:ablation}).

\begin{figure}
\includegraphics[width=\linewidth]{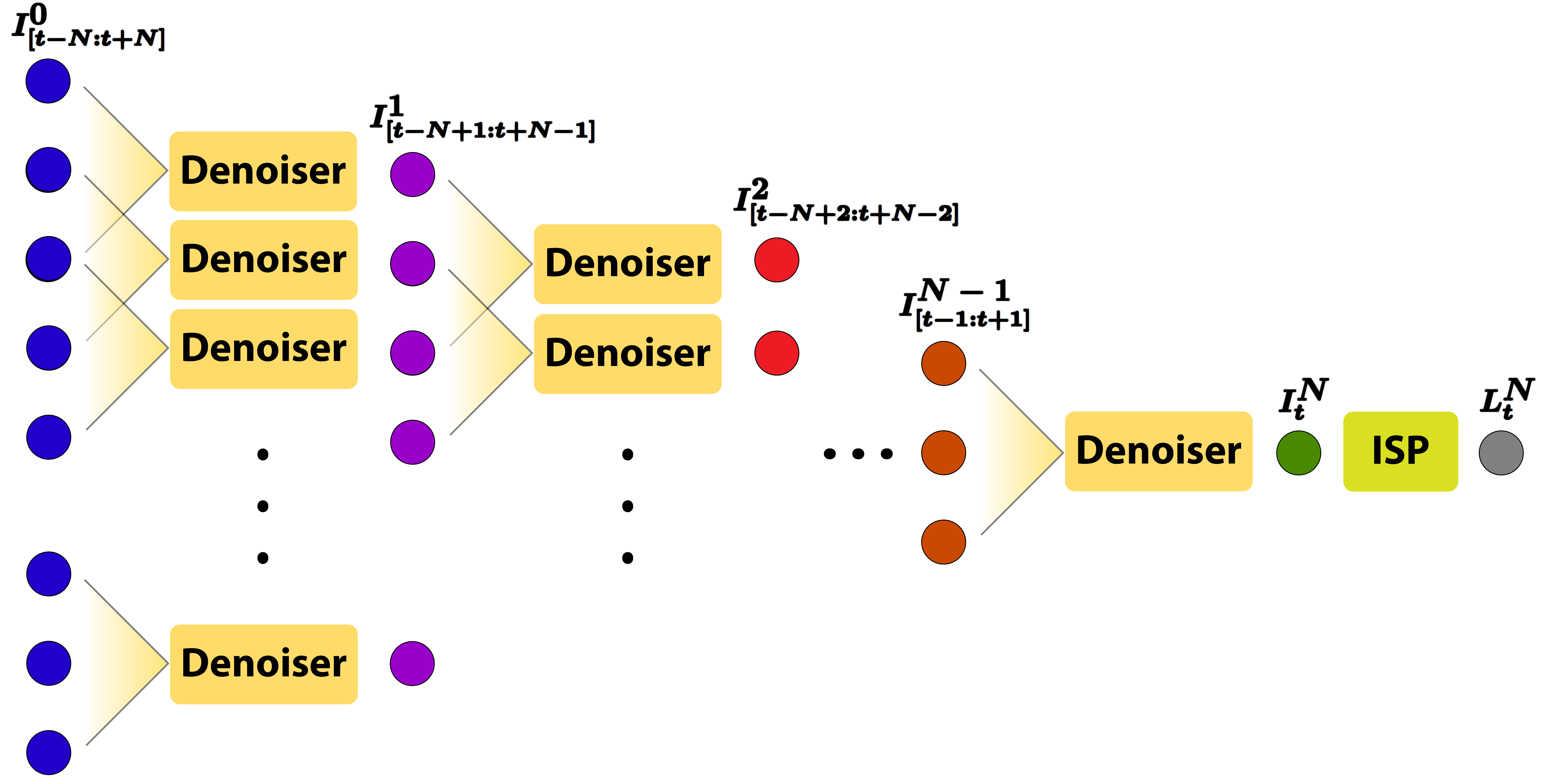}
\caption{The multi-stage denoiser progressively denoises raw noisy sequence $I^{0}_{[t-N:t+N]}$ until we obtain the final denoised raw frame $I^{N}_{t}$. The ISP~\cite{Yue2020} converts the raw frame, $I^{N}_{t}$, to sRGB, $L^{N}_{t}$. Note that, each denoiser in our system, generates the output from three consecutive frames.}
\label{fig:multistage}
\end{figure}

\begin{figure*}
\includegraphics[width=\linewidth]{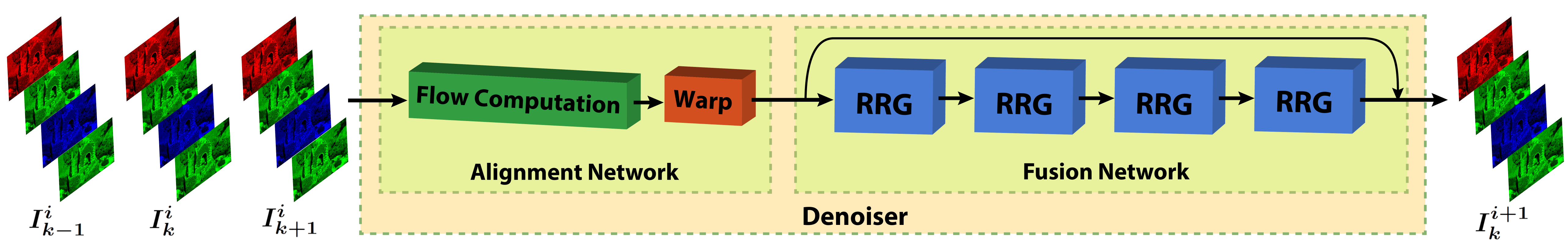}
\caption{Architecture of the denoiser block in the multi-stage denoiser. The two-input flow network separately processes the two adjacent frame pairs, ($I^i_{k-1}$, $I^i_{k}$) and ($I^i_{k+1}$, $I^i_{k}$), and generates a pair of flows which are used to warp the adjacent frames to $I^{i}_{k}$. The fusion network then combines the aligned frames to synthesize the intermediate denoised frame, $I^{i+1}_{k}$.}
\label{fig:denoiser}
\vspace{-0.05in}
\end{figure*}

We address this problem by proposing a multi-stage denoiser which denoises the input sequence by processing a smaller sub-sequence of overlapping inputs at each stage, as shown in Fig.~\ref{fig:multistage}. In our system, the entire sequence is progressively denoised, gradually improving the denoised result until we obtain the final denoised frame. Specifically, at each stage, we use three consecutive frames, $I^{i - 1}_{[k-1:k+1]}$, as input to generate an intermediate denoised frame, $I^{i}_{k}$, where $i$ denotes the stage number. Note that, the noisy input sequence, $I^{n}_{[t-N:t+N]}$, can be considered the output of the zero$^\text{th}$ stage, $I^{0}_{[t-N:t+N]}$. The intermediate frames at each stage, $I^{i}_{[t-N+i:t+N-i]}$, are recursively used as input to the next stage until we obtain the final denoised frame at the $N^{\text{th}}$ stage, $I^{N}_{t}=I^{d}_{t}$. Note that, by performing the denoising process in $N$ stages, the denoised frame is obtained by combining the information of $2N+1$ neighboring frames.

In this paper, we generate the results using a two-stage denoiser and, thus, utilize the information of 5 ($2N+1$ where $N=2$) consecutive frames to reconstruct each denoised frame. The first stage outputs the intermediate denoised sequence $I^{1}_{[t-1:t+1]}$ from the noisy input sequence $I^{0}_{[t-2:t+2]}$. The intermediate sequence with three frames is again fed to the denoiser to obtain the final denoised frame, $I^{d}_{t}$. Our approach can effectively utilize the information of temporally distant frames by performing the alignment and fusion on only three consecutive frames at multiple stages. 

Note that, Tassano et al.~\cite{Tassano2020FastDVDnet} also propose a two-stage denoiser, however, each denoising block in their approach is modeled by a single network. In contrast, we break down the denoising block into two stages of alignment and fusion. By explicitly aligning the neighboring neighboring frames we are able to utilize their information more effectively. Furthermore, unlike Tassano et al., we use denoisers (flow and fusion networks) with shared weights across different stages. By doing so, we reduce the number of trainable parameters and implicitly ensure that the input frames are denoised progressively at multiple stages. Below we discuss the two stages of each denoising block, i.e., alignment and fusion.

\textbf{Alignment}
\quad The goal of alignment is to register the three consecutive input frames by generating a pair of flows, $F_{k\to{k-1}}$ and $F_{k\to{k+1}}$, from the reference frame, $I^{i}_{k}$, to the previous and next frame, $I^{i}_{k-1}$ and $I^{i}_{k+1}$, respectively. The flows are then used to warp the two neighboring frames to produce a set of aligned frames. We use an iterative flow network architecture, proposed by Teed et al.~\cite{Teed2020}, which essentially generates a coarse flow in the first pass and then enhances it in following passes. This flow network calculates a 128 channel contextual map at $1/8$ resolution for the reference frame to assist with flow estimation. We upsample this contextual map by a factor of 8 (using Teed et al.'s learned upsampling module) and pass it to the fusion network so that the network is cognizant of a scene's context such as object boundaries. A flow generated by our network along with a warped image is shown in Fig.~\ref{fig:mask} (left).

\textbf{Fusion}
\quad Here, we merge the output of alignment network, $\tilde{I}^{i}_{[k-1:k+1]}$, with the help of reference frame's contextual information, to generate the denoised frame, $I^{i+1}_{k}$. For fusion, we utilize a residual CNN comprised of recursive residual group (RRG) blocks, proposed by Zamir et al.~\cite{Zamir2020CycleISP}, as shown in Fig.~\ref{fig:denoiser}. The RRG block utilizes attention blocks~\cite{Hu2017Squeeze, Woo2018CBAM} to identify important features, while discarding trivial information like artifacts caused by inaccurate warping.

\subsection{Loss}
Most of the existing denoising techniques use a pixel-wise loss~\cite{Yue2020, brooks2019unprocessing, Zamir2020CycleISP, Yu2019DIDN}, e.g., $\mathcal{L}_{1}$, to train their system. Unfortunately, while a network trained using a pixel-wise loss is able to remove the noise, it does so by blurring out the details. Therefore, we instead propose to train our system with an adversarial loss to reconstruct denoised videos with realistic details, while effectively denoising the smooth regions. Our adversarial loss is defined as follows:

\vspace{-0.1in}
\begin{multline}
\footnotesize
\min_{G}\bigg(\Big(\max_{D} \mathcal{L}_{\text{adv}}(G,D)\Big) + \lambda_{{f}} \mathcal{L}_{\text{feat}}(G,D) \\ 
+ \lambda_{{r}} \mathcal{L}_{\text{recn}}(G) + \lambda_{{p}} \mathcal{L}_{\text{prcp}}(G)\bigg),
\label{eq:adversarialLossEqn}
\end{multline}
\vspace{-0.1in}

\noindent where $D$ and $G$ denote the discriminator and generator, respectively. We use the network proposed by Wang et al. \cite{Wang2018pix2pixHD} as our discriminator and the generator is our multi-stage denoiser. Moreover, $\mathcal{L}_{\text{adv}}$, $\mathcal{L}_{\text{feat}}$, $\mathcal{L}_{\text{recn}}$, and $\mathcal{L}_{\text{prcp}}$ are adversarial, feature, reconstruction, and perceptual losses, respectively. Finally, the corresponding weights are set to $\lambda_{{f}}=1e^{-2}$, $\lambda_{{r}}=1e^{-1}$, and $\lambda_{{p}}=5e^{-3}$. Next, we discuss each loss in detail.

\textbf{Adversarial Loss $\bm{\mathcal{L}_{\text{adv}}}$}
\quad An adversarial loss adds realistic details to the output, thus, giving a natural feel to the synthesized videos. Ideally, we want to enhance the details in the textured areas, while effectively denoising smooth areas. However, by using a na\"ive adversarial loss, our system introduces high-frequency artifacts in the smooth areas of the generated frames. This is mainly because the discriminator in this case, interprets the high-frequency artifacts in the smooth areas as real textures.

\begin{figure}
\includegraphics[width=\linewidth]{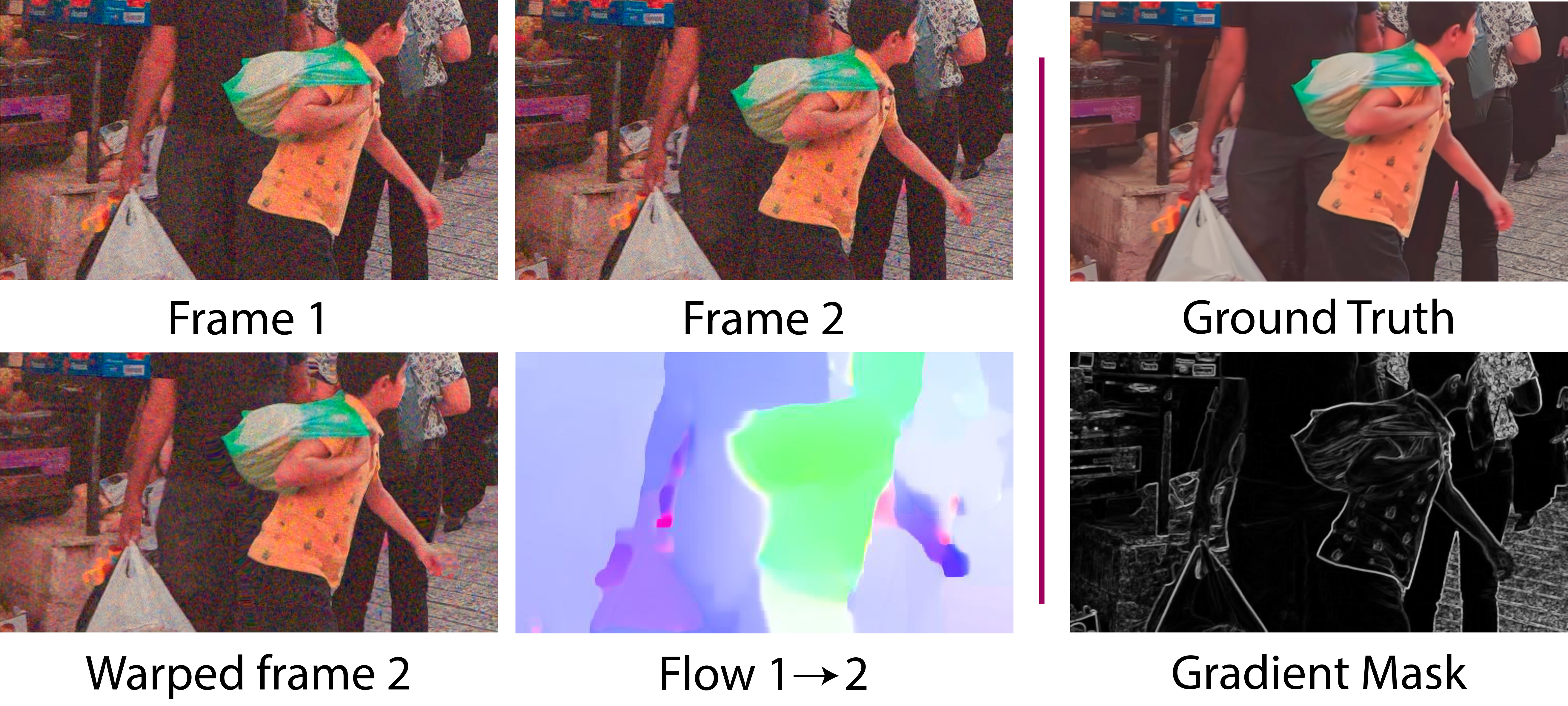}
\caption{On the left, we show two consecutive noisy frames and the estimated flow from frame 1 to 2. We also show the warped version of frame 2 using the estimated flow. Note that, we estimate the flow and perform the warping on raw images, however, we convert the warped image to sRGB domain for visualization. On the right, we show a ground truth frame with its corresponding gradient mask, $f$.}
\label{fig:mask}
\end{figure}

To avoid this issue, we propose to condition the discriminator on a soft mask that identifies the textured areas (see Fig.~\ref{fig:mask}). In this way, the discriminator is able to detect the generated frames with high-frequency artifacts in smooth areas as fake, which consequently encourages the generator to remove such artifacts. 

To obtain the mask, we first compute vertical and horizontal gradients of the ground truth grayscale frame.
We then filter the two gradients using a box filter of size 3, before computing the gradient magnitude, $f'$. Finally, we obtain the gradient mask $f$ by highlighting the gradients as follows:
\begin{equation}
    f = \tanh(f' / \alpha),
\end{equation}
where $\alpha$ is a normalizing parameter and is set to $\alpha=0.8\sqrt{2}$ in our experiments. This map is used along with the fake and real images as inputs to the discriminator, guiding the network to enhance details in the highlighted regions, while effectively denoising the smooth regions. In our system, we use the hinge adversarial loss~\cite{lim2017geometric, Miyato2018spectral, Zhang2018a, Tran2017DeepAH}, defined as follows:
\begin{equation}\label{eq:GAN_Loss}
\begin{split}
L_D = \; & -\mathbb{E} [\min(0, -1 + D(L^{c}_{t}, f))] \\ &- \mathbb{E} [\min(0, -1-D(G(I^{n}_{[t-N:t+N]}), f)) ], \\
L_G = \;  &-\lambda_{{g}}\mathbb{E} [D(G(I^{n}_{[t-N:t+N]}), f)],
\end{split}
\end{equation}
\noindent where $L^{c}_{t}$ is the clean ground truth frame ($L$ denotes frames in sRGB domain), $f$ is the ground truth gradient mask, and $\lambda_g$ is the weight of the generator term which we set to $5e^{-5}$.

\begin{figure*}
\includegraphics[width=\linewidth]{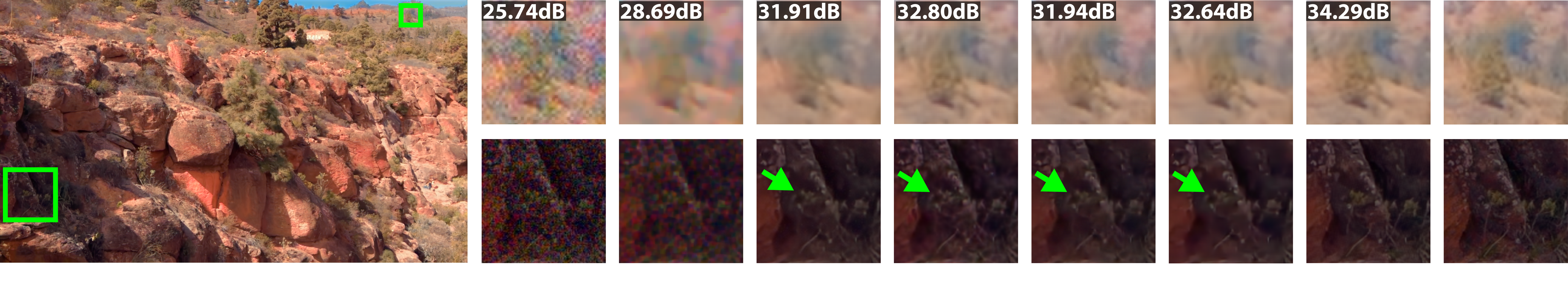}
\includegraphics[width=\linewidth]{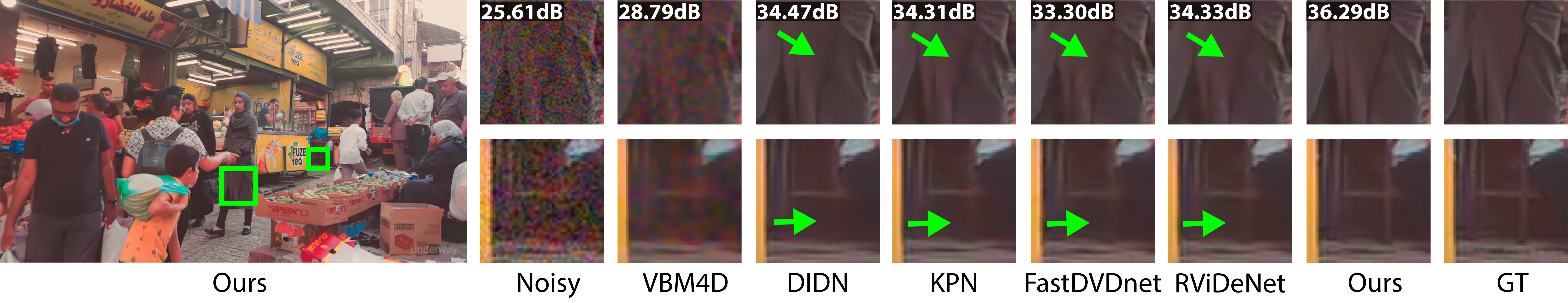}
\caption{Comparison against state-of-the-art image and video denoising methods on \textsc{Canyon} and \textsc{Market} scenes. We use the method of Brooks et al.~\cite{brooks2019unprocessing} to produce the input raw noisy videos. Zoom into the electronic version to see the differences. The videos for DIDN, RViDeNet, and our approach are provided in the supplementary video.}
\label{fig:outdoorsyncomparison}
\vspace{-0.05in}
\end{figure*}

\textbf{Feature Loss $\bm{\mathcal{L}_{\text{feat}}}$}
\quad We use a feature loss proposed by Wang et al.~\cite{Wang2018pix2pixHD} to stabilize the adversarial training. It is defined as:

\vspace{-0.1in}
\begin{equation}
    \mathcal{L}_{\text{feat}} = \sum_{i=1}^T \frac{1}{N_i}\Vert D^{i}(L^{c}_{t}, f) -D^{i}(G(I^{n}_{[t-N:t+N}), f)\Vert_1,
\end{equation}
\vspace{-0.1in}

\noindent where $D^{i}(.)$ and $T$ denote the discriminator's $i^{\text{th}}$-layer's output and total number of discriminator layers, respectively. We use the 4-layer discriminator defined by Wang et al.~\cite{Wang2018pix2pixHD} with spectral normalization~\cite{Miyato2018spectral}.

\textbf{Reconstruction loss $\bm{\mathcal{L}_{\text{recn}}}$}
\quad We use an $L_{1}$ reconstruction loss in both sRGB and raw domains with a weighting parameter, $\delta=0.5$, defined as:

\vspace{-0.1in}
\begin{equation}
    \mathcal{L}_{\text{recn}} = \Vert I^{c}_{t} - I^{d}_{t}\Vert_1 + \delta\Vert L^{c}_{t} - L^{d}_{t}\Vert_1.
\end{equation}
\vspace{-0.1in}

\textbf{Perceptual loss $\bm{\mathcal{L}_{\text{prcp}}}$}
\quad We use a VGG-based perceptual loss, as proposed by Chen and Koltun~\cite{Chen_2017_ImSyn}, to improve the texture in the results. The perceptual loss is defined as: 

\vspace{-0.1in}
\begin{equation}
    \mathcal{L}_{\text{prcp}} = \sum_{i=1}^N\frac{1}{M_i}||V^{(i)}(L^{c}_{t})-V^{(i)}(L^{d}_{t})||_1,
\end{equation}
\vspace{-0.1in}

where $V^{(i)}$ denotes the $i^{\text{th}}$-layer with $M_i$ elements of the VGG network ~\cite{Simonyan2015VGG}.

\definecolor{LightYellow}{rgb}{1,1,0.5}
\begin{table}[!t]
\renewcommand{\arraystretch}{1.3}
\caption{Numerical comparisons on our synthetic test set.}
\centering
\begin{footnotesize}
\begin{tabularx}{\linewidth}{X|cc|ccc}
  \hline\hline
  & \multicolumn{2}{c|}{Raw} & \multicolumn{3}{c}{sRGB}\\
  & PSNR & SSIM & PSNR & SSIM & LPIPS\\
  \hline
  Noisy & 32.53 & 0.811 & 31.82 & 0.829 & 0.2727\\
  VBM4D~\cite{Maggioni2012} & - & - & 33.41 & 0.898 & 0.1982\\
  DIDN~\cite{Yu2019DIDN} & 35.27 & 0.888 & 36.54 & 0.952 & 0.0700\\
  KPN~\cite{mildenhall2018kpn} & 39.24 & 0.967 & 36.85 & 0.959 & 0.0612\\
  FastDVDnet~\cite{Tassano2020FastDVDnet} & - & - & 36.12 & 0.942 & 0.0921\\
  RViDeNet~\cite{Yue2020} & 40.23 & 0.973 & 36.94 & 0.957 & 0.0698\\
  Ours & \textbf{41.09} & \textbf{0.979} & \textbf{37.96} & \textbf{0.967} & \textbf{0.0424}\\
  \hline\hline
\end{tabularx}
\end{footnotesize}
\label{tab:ResultsSynthetic}
\end{table}

\subsection{Training}

Training the two networks in our multi-stage system in an end to end manner is difficult. Therefore, we perform the training in two stages. First, we train the flow network separately to ensure it can effectively align the noisy input images. To perform the training, we use an $\mathcal{L}_1$ warping loss, $\mathcal{L}_w$, between clean sRGB reference and warped adjacent frames in addition to a total variation smoothing loss, $\mathcal{L}_{\text{tv}}$. We apply the combination of warping and TV losses to each iterations using the weighting scheme similar to Teed. et al~\cite{Teed2020}:

\vspace{-0.1in}
\begin{equation}
\mathcal{L}_{\text{flow}} = \sum_{i=1}^{N} \gamma^{i-N}(\alpha \mathcal{L}^{i}_{w} + \mathcal{L}^{i}_{tv}),
\end{equation}
\vspace{-0.1in}

\noindent where $\gamma=0.8$ and $\alpha=100$. Here, $\gamma^{i-N}$ increasingly weights the latter flows of the iterative network to ensure effective coarse to fine flow calculation. Once the flow network is trained, we use it in our multi-stage denoiser and train the fusion network by minimizing the loss in Eq.~\ref{eq:adversarialLossEqn}. 

\highlighttext{Currently, there is a lack of large scale high-quality FHD video dataset that can be used to train video restoration techniques. Therefore, we use youtube-dl to download 200 4K resolution videos from various YouTube channels. Most of the 4K YouTube videos are made up of smaller clips pieced together. From these, we extract around 500 high-quality clips with good lighting and downsample them to 1080p to mitigate the noise and compression artifacts. Most of the videos are footage captured by enthusiasts using drones or personal cameras. Therefore, our dataset has a variety of scenes with urban/rural landscape containing rigid (e.g., vehicles) and non-rigid (e.g., humans, animals, etc.) motion, as well as diverse camera motion.} Next, we extract 16,600 training samples each containing 8 consecutive frames. We augment the data using random horizontal and vertical flipping, and random change in input video sequence direction. \highlighttext{This dataset is crucial for learning a powerfull model with the ability to generalize to a variety of scenes with different conditions.}

To train our flow network, we randomly choose two frames from the sample and randomly crop them to patches of size $384\times384$. To further augment the data, we apply PyTorch's color jitter with parameters ($\text{hue}=0.2/\pi$, $\text{contrast}=0.2$, $\text{brightness}=0.2$, $\text{saturation}=0.2$). The jitter is applied to the two frames separately with a probability of 0.2. Otherwise the same jitter augmentation is applied to both the frames. For training our fusion network in our multi-stage denoiser, we use patches of size $96\times96$ from 5 consecutive frames in our training samples.

\begin{figure*}
\includegraphics[width=\linewidth]{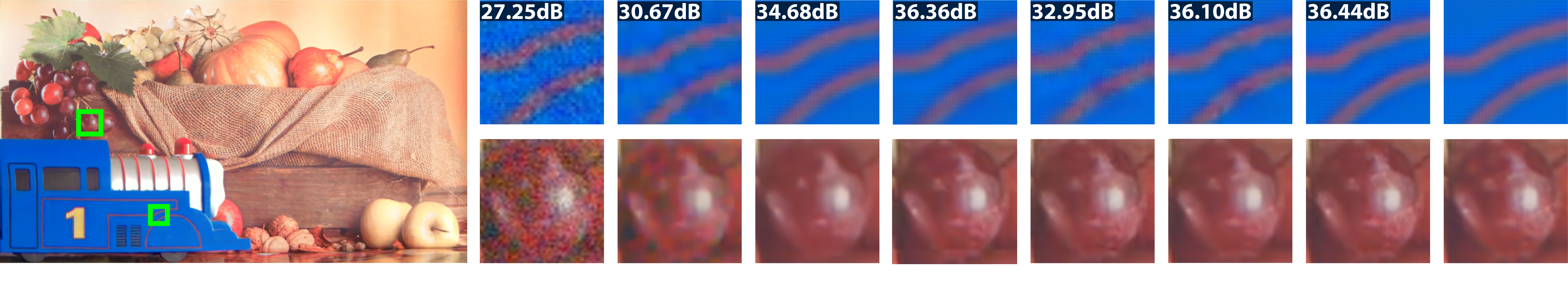}
\includegraphics[width=\linewidth]{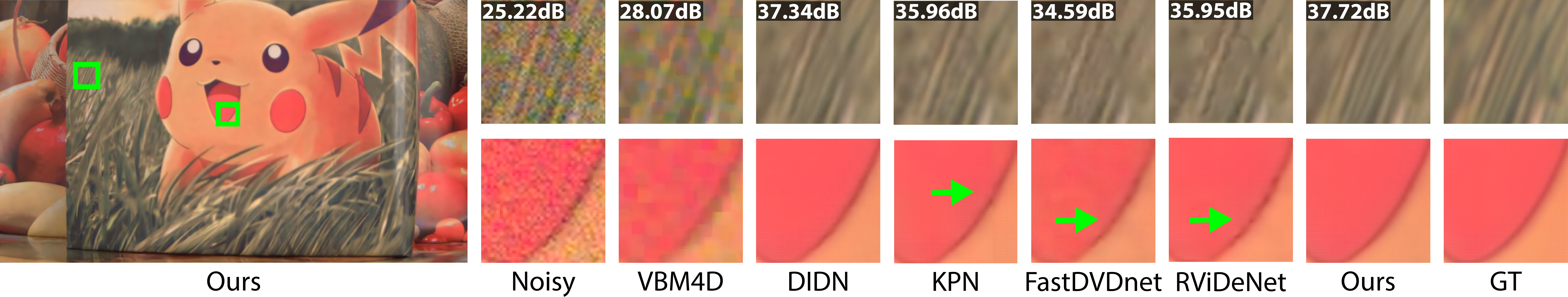}
\caption{Comparison against state-of-the-art image and video denoising methods on real indoor scenes \textsc{Train} and \textsc{Pikachu} by Yue at al.~\cite{Yue2020}.}
\label{fig:indoorcomparison}
\vspace{-0.05in}
\end{figure*}

The synthetic training dataset is composed of sRGB frames but our goal is to train a network to denoise raw video sequences. Therefore, to generate realistic raw noisy data, we use the unprocessing pipeline proposed by Brooks et al.\cite{brooks2019unprocessing} to convert the sRGB frames to raw. We then add shot and read noise to the raw frames using Poisson $\mathcal{P}$ and $\mathcal{N}$ Gaussian distribution~\cite{brooks2019unprocessing, mildenhall2018kpn, Yue2020} as follows:

\vspace{-0.1in}
\begin{equation}
    x_p \sim \sigma_s^2\mathcal{P}(y_p/\sigma_s^2) + \mathcal{N}(0,\sigma_r^2),
\end{equation}
\vspace{-0.1in}

\noindent where $x_p$ is the noisy observation and $y_p$ is the true intensity at pixel $p$. Moreover, $\sigma_r$ and $\sigma_s$ are read and shot noise parameters. We use the same set of shot and read parameters as RViDeNet~\cite{Yue2020} to be able to fairly compare with their approach on synthetic and real test scenes.

%% file: Results.tex
\begin{table}[!t]
\renewcommand{\arraystretch}{1.3}
\caption{Results on real indoor test set by Yue et al.~\cite{Yue2020}.}
\centering
\begin{footnotesize}
\begin{tabularx}{\linewidth}{X|cc|ccc}
  \hline\hline
  & \multicolumn{2}{c|}{Raw} & \multicolumn{3}{c}{sRGB}\\
  & PSNR & SSIM & PSNR & SSIM & LPIPS\\
  \hline
  Noisy & 31.99 & 0.733 & 31.72 & 0.751 & 0.4688\\
  VBM4D~\cite{Maggioni2012} & - & - & 34.46 & 0.908 & 0.2076\\
  DIDN~\cite{Yu2019DIDN} & 35.67 & 0.853 & 39.72 & 0.977 & 0.0477\\
  KPN~\cite{mildenhall2018kpn} & 43.06 & 0.986 & 39.77 & 0.979 & 0.0443\\
  FastDVDnet~\cite{Tassano2020FastDVDnet} & - & - & 37.45 & 0.958 & 0.0791\\
  RViDeNet~\cite{Yue2020} & 43.63 & 0.987 & 39.74 & 0.978 & 0.0451\\
  Ours & \textbf{43.96} & \textbf{0.988} & \textbf{40.40} & \textbf{0.981} & \textbf{0.0357}\\
  \hline\hline
\end{tabularx}
\label{tab:ResultsReal}
\end{footnotesize}
\end{table}

\begin{figure*}
\includegraphics[width=\linewidth]{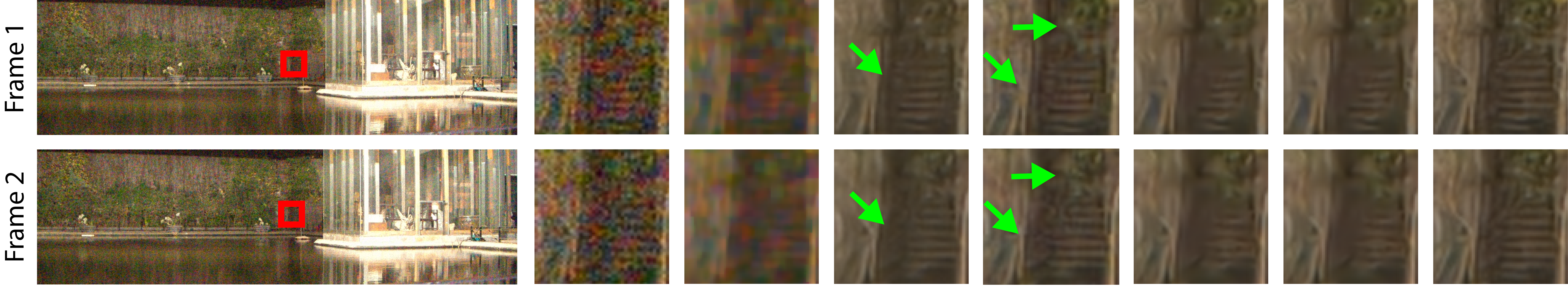}
\includegraphics[width=\linewidth]{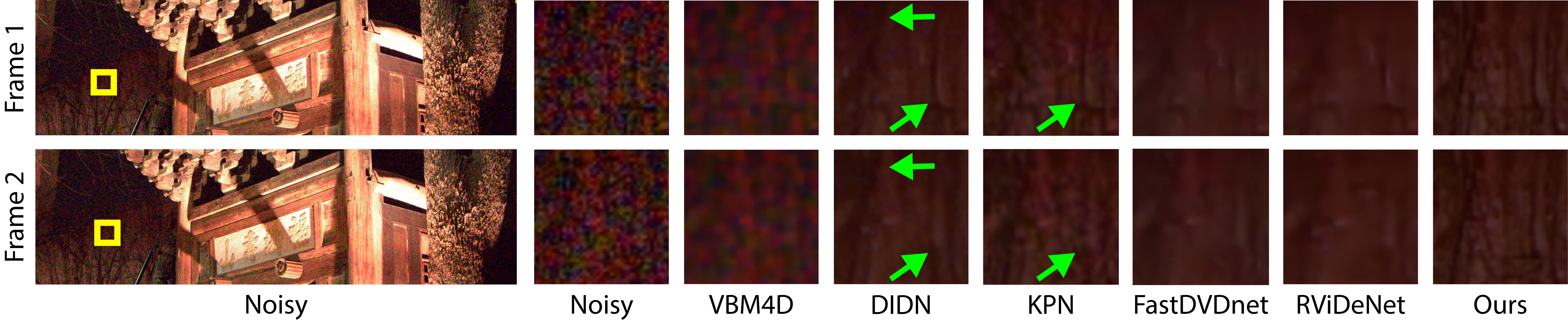}
\caption{Comparison against state-of-the-art image and video denoising methods on real outdoor scenes \textsc{Pool} and \textsc{Pavilion} by Yue at al.~\cite{Yue2020}. For each scene, we show the results on two consecutive frames. Note that, while DIDN reconstructs some of the details, they are not consistent across consecutive frames, as indicated by the green arrows.}
\label{fig:outdoorrealcomparison}
\vspace{-0.05in}
\end{figure*}

\section{Results}
We implemented our model in PyTorch and used Adam~\cite{Kingma15Adam} for training. We train the alignment network for 400k iterations using a learning rate of 3e-4, optimizer parameters, $\beta_1=0.9$ and $\beta_2=0.999$, and batch size of 5. For adversarial training, we use the two time-scale update rule (TTUR) by Heusel et al.~\cite{Heusel2017ttur} for better convergence. The generator and discriminator learning rates are set to 1e-4 and 4e-4, respectively, with optimizer parameters, $\beta_1=0.0$ and $\beta_2=0.9$, and batch size of 4. We decrease the learning rates by a factor of 10 after 200k iterations and train it for a total of 400k iterations which takes 5 days on a RTX 2080 Ti GPU. \highlighttext{The source code along with our synthetic \textit{test} set is available at} \url{https://github.com/avinashpaliwal/MaskDnGAN}.

We compare against image denoising approach of Yu et al.~\cite{Yu2019DIDN} (DIDN), \highlighttext{burst image denoising approach of Mildenhall et al.~\cite{mildenhall2018kpn} (KPN)} and video denoising methods by Maggioni et al.~\cite{Maggioni2012} (VBM4D), Tassano et al.~\cite{Tassano2020FastDVDnet} (FastDVDnet) and Yue et al.~\cite{Yue2020} (RViDeNet). For all the approaches, we use the source codes provided publicly. Moreover, for fair comparisons, we retrain DIDN, \highlighttext{KPN}, FastDVDnet, and RViDeNet with all their losses on our synthetic dataset. \highlighttext{We use a kernel size of $K=15$ for KPN.} Finally, we set the number of input frames in \highlighttext{KPN and} RViDeNet to 5 during retraining to \highlighttext{have the same input as ours}. Note that, we do not provide the noise level information as input to \highlighttext{KPN and} FastDVDnet since none of the other approaches, including ours, utilize such information. We obtain the $\sigma$ parameter for VBM4D using a grid search to achieve the best PSNR on the test sets.

\subsection{Synthetic Videos}
We create a synthetic test set of 12 high-quality 1080p videos extracted from 4K YouTube videos under Creative Commons license. Table.~\ref{tab:ResultsSynthetic} shows the quantitative comparison of our system against other state-of-the-art approaches in terms of PSNR, SSIM~\cite{Wang2004ssim}, and LPIPS\cite{Zhang2018lpips}. Our adversarial multi-stage approach outperforms the other methods, especially in terms of perceptual quality, as demonstrated by the LPIPS metric.

Next, we compare our approach against the other methods on two of these synthetic scenes in Fig.~\ref{fig:outdoorsyncomparison}. In both scenes, VBM4D is unable to effectively denoise the videos. In the \textsc{Canyon} scene (top), the camera is moving backwards and, thus, the bottom areas have large motions, while the top regions (horizon) remain relatively static. FastDVDnet and RViDeNet are able to align objects in the top regions and generate sharper results in comparison to DIDN, a single image denoising approach. On the other hand, both FastDVDnet and RViDeNet suffer in regions with significant motion and generate slightly blurrier results compared to DIDN. \highlighttext{Although KPN produces sharper results than FastDVDnet and RViDeNet, it is not able to reproduce accurate color and texture in very noisy regions}. Our method is able to generate sharper results \highlighttext{with consistent color} in all regions. The \textsc{Market} scene (bottom) contains both object and camera motions. Other approaches are unable to generate sharp details, while our method is able to recover the texture on the moving objects.

\subsection{Real dataset}
Here, we show comparisons on the real test set provided by Yue et al.~\cite{Yue2020} which is divided into 5 indoor and 10 outdoor scenes. The indoor scenes are captured in a controlled environment with ground truth frames, while the outdoor scenes are captured without ground truth. Both test sets consist of five ISO levels, covering a variety of different noise levels. Each indoor scene is a video sequence of 7 frames and we compute the quantitative results only on the valid frames, i.e., frames 3 to 5. As shown in Table.~\ref{tab:ResultsReal}, our approach significantly outperforms the other methods according to all metrics in both sRGB and raw domains.

We show comparisons against other approaches on two scenes from the indoor test set in Fig.~\ref{fig:indoorcomparison}. The \textsc{Train} scene (top) has a static background with the blue train moving horizontally. VBM4D denoises the train but is unable to reproduce the background texture. DIDN performs well in the smooth regions of the train, but blurs the details of the textured areas like the grapes. FastDVDnet, and RViDeNet are unable to align the adjacent frames and generate artifacts on the moving train, but are able to combine frames in static regions (e.g., grapes) to reconstruct the details. Furthermore, FastDVDnet generates artifacts in saturated regions (see supplementary pdf). \highlighttext{Here, KPN produces comparable result to ours as it can effectively combine the images in static regions (top inset) and denoise the smooth areas on the train}.  In the \textsc{Pikachu} scene (bottom), the box at the center with Pikachu is being horizontally rotated while the background is static. VBM4D retains some noise in the form of color artifacts. DIDN removes the noise, but also blurs out the details. KPN cannot effectively combine the images and produces artifacts on the textured regions (top inset), but effectively denoises the smooth regions (bottom inset). FastDVDnet, and RViDeNet have alignment artifacts as seen in both the textured areas and smooth regions around the mouth. In comparison, our method produces results without visible artifacts and reconstructs the details by effectively combining all the input frames.


Next, we compare our approach against the other methods on two outdoor scenes by showing the results on two consecutive frames (Fig.~\ref{fig:outdoorrealcomparison}). VBM4D is not only unable to reconstruct the details, but also preserves some of the input noise. DIDN produces results with more details compared to FastDVDnet and RViDeNet, but the textures change in the consecutive frames (see the green arrows) resulting in flickering artifacts. \highlighttext{Similarly, KPN generates sharper frames but the texture changes between consecutive frames, especially in the \textsc{Pavilion} scene where the colors and texture vary drastically.} FastDVDnet and RViDeNet mitigate the flickering but also smooth out the details in the process. Our result has more details, while being consistent across the two frames (see supplementary video).

\subsection{\highlighttext{Temporal Consistency}}
\highlighttext{We numerically evaluate the temporal consistency of our approach against the other approaches in Table~\ref{tab:TemporalConsistency}, in addition to the visual comparison shown in Fig.~\ref{fig:outdoorrealcomparison}. We use a flow based warping error defined by Lai et al.~\cite{Lai2018TempCons}. Here, we use the flows computed on clean frames to warp the adjacent denoised frames, $L_t^d$ and $L_{t+1}^d$, to the current frame. The average warping error (E$_{\text{w}}$) is then computed by averaging the masked $\mathcal{L}_1$ error between the warped adjacent frames. In terms of temporal consistency, our approach outperforms the other approaches by a large margin while generating sharper results.}

\subsection{Inference Performance}
We compare the inference time of our approach against the other approaches in Table~\ref{tab:InferencePerf}. While DIDN, \highlighttext{KPN} and FastDVDNet are faster than our approach, they produce results with significantly lower quality. Compared to RViDeNet, our method is considerably faster and produces better results. The majority of timing in our approach is spent during flow computations. Therefore, we can cut down our timing to 1.25s by reusing the flows computed in the first stage.


\begin{table}[!t]
\renewcommand{\arraystretch}{1.3}
\centering
\caption{Temporal consistency evaluation using average warping error~\cite{Lai2018TempCons} on the synthetic test set.}
\begin{footnotesize}
\begin{tabularx}{\linewidth}{Xccccc}
  \hline\hline
  & DIDN & KPN & FastDVDnet & RViDeNet & Ours\\
  \hline
   E$_{\text{w}}$ (x 10$^{-4}$) \qquad & 8.81 & 7.99 & 8.34 & 7.96 & \textbf{7.02}\\
  \hline\hline
\end{tabularx}
\label{tab:TemporalConsistency}
\end{footnotesize}
\end{table}

\newcolumntype{Y}{>{\centering\arraybackslash}X}
\begin{table}[!t]
\renewcommand{\arraystretch}{1.3}
\caption{Comparison of inference time for generating a 1080p frame on RTX 2080 Ti.}
\centering
\begin{footnotesize}
\begin{tabularx}{\linewidth}{cYcccY}
  \hline\hline
  VBM4D & DIDN & \highlighttext{KPN} & FastDVDNet & RViDeNet & Ours\\
  \hline
   124.90s & 1.20s & \highlighttext{0.89s} & 0.74s & 9.82s & 1.47s\\
  \hline\hline
\end{tabularx}
\label{tab:InferencePerf}
\end{footnotesize}
\end{table}

\subsection{Ablation Study}
\highlighttext{We demonstrate the effect of each component in our model on the denoising quality in Tables~\ref{tab:AblationComponent}~and~\ref{tab:TemporalConsistencyAblation}. Increasing the number of inputs from 3 to 5, improves the performance in terms of both, image quality and temporal consistency, showing that our system effectively combines information from the additional frames. Switching to the 2-stage model (Ours5M) further improves performance and temporal consistency, thereby demonstrating the effectiveness of the multi stage architecture. As seen, going beyond two stages to three stages (Ours7M) reduces the performance as training such system is difficult. Next we show that while training the system on our full loss, including the adversarial term, improves the perceptual quality (LPIPS), it lowers the PSNR and SSIM due to poor performance in smooth regions (high frequency artifacts). Conditioning the discriminator with the gradient mask gives the best PSNR and SSIM, while further improving the perceptual quality.}

\highlighttext{Finally, we show the inference time for different number of stages in Table~\ref{tab:inferencetime}. While the number of networks used to generate a single denoised frame increase exponentially with the number of stages (3+1 for 2-stage; 5+3+1 for 3-stages), the amortized cost of inference increases linearly with the number of stages, as shown in Table~~\ref{tab:inferencetime}. This is because most of the computation from the previous frame can be reused in computation of the current denoised frame.}

\begin{table}[!t]
\renewcommand{\arraystretch}{1.3}
\caption{Evaluating the effect of components in our model on the denoising quality.}
\centering
\begin{footnotesize}
\begin{tabularx}{\linewidth}{X|cc|ccc}
  \hline\hline
  \multirow{2}{*}{\qquad Component} & \multicolumn{2}{c|}{Raw} & \multicolumn{3}{c}{sRGB}\\
  &  PSNR & SSIM &  PSNR & SSIM & LPIPS\\
  \hline
   3-Input (1-stage) & 40.87 & 0.978 & 37.67 & 0.965 & 0.0432\\
   5-input (1-stage) & 40.97 & \textbf{0.979} & 37.84 & 0.967 & 0.0423\\
   5-input (2-stage) & \textbf{41.09} & \textbf{0.979} & \textbf{37.96} & \textbf{0.967} & 0.0424\\
   7-input (3-stage) & 40.96 & 0.978 & 37.83 & 0.967 & \textbf{0.0418}\\ \hline
   5-input (2-stage)  & 40.89 & 0.978 & 37.79 & 0.966 & 0.0542\\
   \enskip+ Adversarial & 40.58 & 0.977 & 37.44 & 0.964 & 0.0439\\
   \enskip+ Gradient Mask & \textbf{41.09} & \textbf{0.979} & \textbf{37.96} & \textbf{0.967} & \textbf{0.0424}\\
  \hline\hline
\end{tabularx}
\label{tab:AblationComponent}
\end{footnotesize}
\end{table}

\begin{table}[!t]
\renewcommand{\arraystretch}{1.3}
\centering
\caption{Evaluating the effect of components in our model on the temporal consistency.}
\begin{footnotesize}
\begin{tabularx}{\linewidth}{lYYYY}
  \hline\hline
  & Ours3 & Ours5 & Ours5M & Ours7M\\
  \hline
  E$_{\text{w}}$ (x 10$^{-4}$) \qquad & 7.37 & 7.07 & \textbf{7.02} & 7.22\\
  \hline\hline
\end{tabularx}
\label{tab:TemporalConsistencyAblation}
\end{footnotesize}
\end{table}

\begin{table}[!t]
\renewcommand{\arraystretch}{1.3}
\centering
\caption{Inference time (seconds) for multi-stage architecture per frame.}
\begin{footnotesize}
\begin{tabularx}{\linewidth}{lYYYYY}
  \hline\hline
  N-stage model & 1 & 2 & 3 & 4 & 5\\
  \hline
  inference time (s) & 0.84 & 1.47 & 2.30 & 3.04 & 5.18\\
  \hline\hline
\end{tabularx}
\label{tab:inferencetime}
\end{footnotesize}
\end{table}


\begin{figure}
\includegraphics[width=\linewidth]{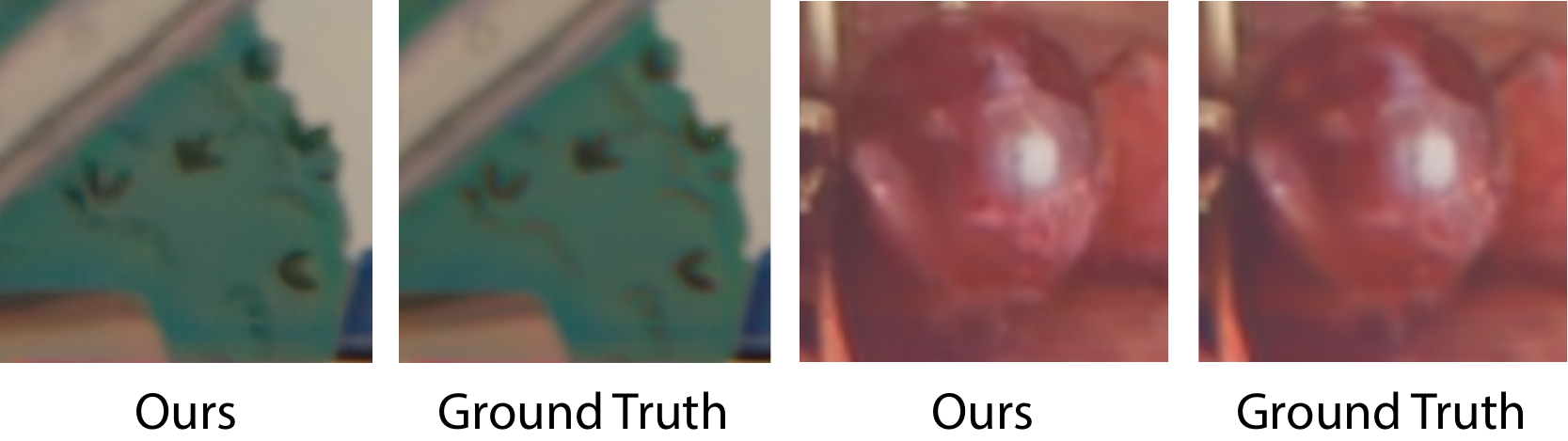}
\caption{On the left, we show an example where our system produces sharper image than the ground truth. On the right, we show a case where our method hallucinate details that do not exist in the ground truth image. Nevertheless, our results are still visually pleasing.}
\label{fig:oversharpening}
\end{figure}

\subsection{Limitations}
Our approach has some limitations. First, our model is unable to effectively denoise videos with noise levels outside the training noise distribution, as shown in Fig.~\ref{fig:outnoise}. Nevertheless, our approach still produces better results than DIDN and RViDeNet, demonstrating better generalization ability of our method. Moreover, in some cases (see Fig.~\ref{fig:oversharpening}), our adversarial system oversharpens the images (left) or hallucinate details (right) that do not exist in the ground truth image. However, this is not a major issue as our results are visually plausible. Finally, compared to single-stage, our multi-stage approach is more computationally demanding, especially as the number of stages increase.

%% file: Conclusion.tex
\section{Conclusion \& Future Work}
We present a novel deep learning approach for raw video denoising. We propose a multi-stage denoiser with a denoiser block further divided into alignment and fusion stages. The proposed denoiser alleviates the need to directly align temporally distant frames in the noisy video sequence. We further propose an adversarial training strategy where we condition the discriminator on a gradient mask. Our adversarial approach effectively removes noise, while preserving the details in the scene. We demonstrate that our approach significantly outperforms state-of-the-art methods on both synthetic and real scenes, both visually and numerically in terms of several quantitative metrics.
\highlighttext{In the future, it would be interesting to explore the possibility of using our system for the other video enhancement tasks like video super resolution.}

\begin{figure}
\includegraphics[width=\linewidth]{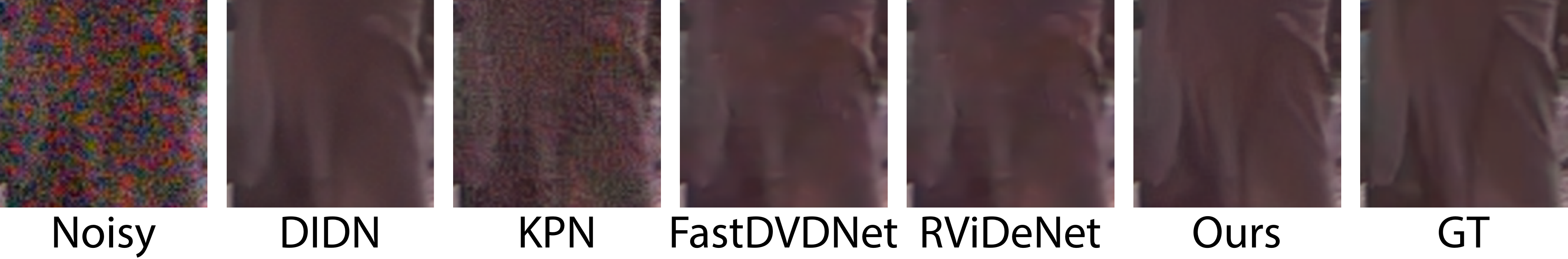}
\caption{For noise outside the training distribution (\texttt{ISO} 38400 in this case), our model introduces artifacts in the smooth regions. However, our result is still much better than other approaches at extracting the underlying texture.}
\label{fig:outnoise}
\end{figure}